# A DYNAMIC BATTERY STATE-OF-HEALTH FORECASTING MODEL FOR ELECTRIC TRUCKS: LI-ION BATTERIES CASE-STUDY

Matti Huotari[1], Shashank Arora[1], Avleen Malhi[1], Kary Främling[2,1]

[1]Aalto University, Espoo, Finland
[2]Umeå University, Umeå, Sweden

## ABSTRACT

*It is of extreme importance to monitor and manage the battery health to enhance the performance and decrease the maintenance cost of operating electric vehicles. This paper concerns the machine-learning-enabled state-of-health (SoH) prognosis for Li-ion batteries in electric trucks, where they are used as energy sources. The paper proposes methods to calculate SoH and cycle life for the battery packs. We propose autoregressive integrated modeling average (ARIMA) and supervised learning (bagging with decision tree as the base estimator; BAG) for forecasting the battery SoH in order to maximize the battery availability for forklift operations. As the use of data-driven methods for battery prognostics is increasing, we demonstrate the capabilities of ARIMA and under circumstances when there is little prior information available about the batteries. For this work, we had a unique data set of 31 lithium-ion battery packs from forklifts in commercial operations. On the one hand, results indicate that the developed ARIMA model provided relevant tools to analyze the data from several batteries. On the other hand, BAG model results suggest that the developed supervised learning model using decision trees as base estimator yields better forecast accuracy in the presence of large variation in data for one battery.*

## 1. INTRODUCTION

The efficient transportation system can improve the flow of goods and diminish the amount of energy used. In this regard, electric vehicles (EVs) have gained attention owing to their effectiveness in reducing oil demands and gas emissions especially in the area of forklifts. These forklifts are not only energy-efficient but are also safer for the drivers as electrically powered forklifts produce no fumes, vibrate less and are quieter than the combustion engine powered forklift trucks. Of the electric components of an EV, the battery is considered as the major bottleneck. To power these EVs, a widely employed battery type is lithium-ion battery. For lithium-ion batteries, the battery state of health (SoH) estimation is used for monitoring and for controlling EV batteries in order to ensure their safe application and to maximize the battery availability for forklift operations.

Voltages available through state-of-the-art single cells are not sufficient for supporting an electric driveline. Therefore, many cells have to be combined in series and in parallel to build up battery packs that are then used as energy sources in EVs. Cells in a battery pack age differently than they would in an isolated environment which affects the service life of the battery packs, and as a result, EVs. This research focuses on EV battery packs in contrast to major studies related to individual battery cells aging [1-3]. We have a unique data set from a lithium-ion battery manufacturer in the EU (European Union); the data set consist of 31 three-year time series of batteries used in forklifts.

Each dataset consists of basic signals of a battery pack. The relationship between these basic signals and the SoH is complex under real conditions [4]. Customarily, the SoH forecasting has relied on equivalent-circuit models; however, more recently statistical techniques and machine-learning techniques have been proposed including ARIMA based statistical method [5], neural networks [5–6], Gaussian processes [7–8], support vector machines [9–10] and ensemble machine learning methods [11–12]. The success of these works shows the capabilities of such an approach. We have also considered how the SoH values have been estimated as presented in the references [13–14]. In the selection of our data-based approaches, as we have a set of batteries to analyze, a key to the selected method is that it is applicable to all of the batteries, preferably in such a way that the results can be compared.

Overall, for data-based approaches, there is a tradeoff between complex hypothesis that fit the training data well, and simpler hypothesis that may generalize better [15]. As for well-generalizing method, making battery related forecasts with ARIMA remains difficult [10]; however, ARIMA provides tools for data analysis over a set of batteries as well, such as Wilcoxon significance test to compare the means of the prediction results [18]. The comparison is relevant in this on surface similar set of batteries, as the working environment for forklifts vary. Furthermore, it has been among the best performing methods in the open M3 and M4 forecast competitions [16–17].; for these reasons, we are using ARIMA.

The primary concern of this paper is to predict future values of SoH. We assume that combination of accurate enough basic signals is available for us so that we can make this prediction. We use ARIMA to build a simple model; furthermore, as in the field of batteries supervised learning techniques have been successfully used to model arbitrarily complex lithium-ion cells [6–12]. This paper also proposes a supervised learning method, bootstrap aggregation (bagging) with decision trees as the base regression estimator, for making forecasts. The bagging method is verified by comparing it to several regression methods: Classification and Regression Trees (CART), Random Forest (RF) and Gradient Boosting Machines (GBM). The performance of all proposed models is assessed based on loss functions, which are root mean square error (RMSE) and explained variance ($EVAR$) for both ARIMA and supervised learning methods. We use different proportions of training data while building models with the goal of generalizing the developed model well to any unseen data in the domain of battery SoH.

The rest of this paper describes how to define SoH and charging cycles from the battery time series in section 2.1. Then it introduces the modeling methods: ARIMA and supervised learning method, bagging in sections 2.2–2.3, followed by the evaluation metrics used in model development in section 2.4. Followingly, we introduce the SoH and cycle life model results in section 3.1. Then in section 3.2, we propose an ARIMA model for forecasting the SoH of the batteries; the developed ARIMA model is then widened to all the 31 batteries utilizing Wilcoxon significance test in section 3.3. In section 3.4, we develop a SoH forecast model with the bagging method. In section 3.5, the paper discusses the developed SoH forecasting models and the key findings. Conclusions and future work are in section 4.

## 2. MATERIALS AND METHODS

### 2.1 Battery data and calculation of derived parameters.

The time series data of lithium-ion batteries from 31 EV forklifts was used for as basis for this study. The data was collected around three years' time from forklifts on different countries and continents using one manufacturer's batteries. The data was collected using the battery manufacturer's sensors that were attached to the batteries and that send their measurement data to a local hub. This data was saved into a nested csv-data file structure according to the battery serial number and the date of the data collected. In this work, we compiled one battery's data to one csv file in order to make the data usable for machine learning usage. At the same time, the transient failures, where the sensor decided to send occasionally nonsense, and other data failures were cleaned, which lead to discarding around ¼ of the overall data at this first stage.

Each dataset consists of basic signals of a battery pack: voltage, current, battery state of charge (SOC) and forklift operating temperature over time (Table 1).

**TABLE 1:** Basic battery signals measured.

| Basic signals | Units |
|---|---|
| Time stamp of the data | 1 min interval |
| Serial number of the battery | - |
| Measured voltage | V |
| Measured current | A |
| SOC | % |
| Ambient temperature | °C |

Among battery manufacturers, there are some other remote battery monitoring systems (BMS) [19]; however, the manufacturer that provided us the battery time series from batteries with the same nominal capacity used in forklifts. This provided us with a reasonably stable environment for pursuing our main target: SoH forecasts over a set of batteries. Albeit the exact data of cell manufacturers was not made public to us, the focus of this study was on battery pack, and overall this manufacturer answered our oral questions promptly in the initial phase of this study. These were the reasons that we selected the battery-pack data as the starting point for our study.

From these basic signals of each battery, we calculated the total energy charged and individual charging pulses that caused the battery's SOC value to increase. This happened at charging pulses over 3 volts and over 5 minutes; however, the few pulses longer than 30 minutes were discarded as manual analysis confirmed that these long pulses to a substantial part contained transient failures.

For each battery, the energy, cumulated energy, pack capacity, and the number of current cycles were calculated using the following equations:

$$E_j = \int i \cdot v \, dt \quad (1)$$

$$\sum_t E = \sum_0^{present} E_j \quad (2)$$

$$C_n = \sum_t E_{\Delta SOC5\%} \cdot \frac{100}{5} \quad (3)$$

Where $i$ is battery current, $v$ is battery voltage, $E_j$ is energy value that is saved every minute, $C_n$ is pack capacity at specific time.

The pack capacity $C_n$ was defined by calculating the energy at selected pulses where the SOC difference was 5% between the start and end of the charging pulse. Furthermore, the SOC values for the selected pulses were between 20–60% as the battery behavior was identified to be stable between that range. The SoH was defined as the ratio of present capacity $C_n$ value to initial capacity value $C_0$ (Eq. 3) under similar conditions as above, i.e. using selected pulses where the SOC difference was 5% between the start and end of the charging pulse and the nominal SOC values were between 20–60%. To calculate initial capacity, we look at the overall trend and locate the point at which capacity starts to increase. $C_0$ is taken as average of all capacities measured under the defined conditions up to that point. Averaging allows to account for the effect of intercellular variations and temperature on battery pack's capacity.

$$SoH = \frac{C_n}{C_0} \quad (4)$$

The ambient temperature for e.g. the battery A was between 19–28 ºC (Fig. 1). The range of ambient temperature for all batteries vary from -20 ºC to 36 ºC. Noteworthy is that only one battery out of 31 had ambient temperature below zero for a longer period, which may have had an adverse effect on the SoH of the battery [20]. On the other hand, there are also batteries with constantly relatively high ambient temperatures (>32 ºC), which also has adverse effect on the SoH [1,21-22]. Moreover, the ambient temperature change showed some seasonality, and this in turn indicated that time series with at least few more 12-month periods are needed in order to build seasonal ARIMA (SARIMA) [16], or in general in order to identify and mitigate the seasonality effects in confirmer manner.

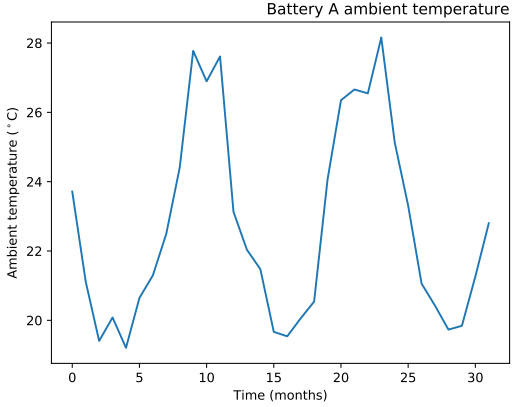

**FIGURE 1:** Ambient temperature for battery A.

## 2.2 ARIMA method

In the field of time series analysis techniques, ARIMA, or autoregressive (AR), integrated (I), modeling average (MA) is employed to learn the suitable ARIMA model from the input time series $Y$ to the estimator $\hat{Y}$ for building a forecast model.

If the timeseries' statistical properties (e.g. the mean and the variance) are not constant in time, the timeseries is made stationary by data transformation. A simple transformation is differencing that is done by the following equation; however, here it must be emphasized that the applicable data transformation method is data-depended:

$$y_t = Y_t - Y_{t-1} \qquad (5)$$

where $y$ denotes the differenced, stationary time series. From this stationary timeseries, the stationary estimator values, $\hat{y}_t$, are calculated by the following equation:

$$\hat{y}_t = \mu + \phi_1 y_{t-1} + \cdots + \phi_p y_{t-p} + \theta_1 e_{t-1} \ldots + \theta_q e_{t-q} \qquad (6)$$

where $\mu$ is constant, $\phi_i y_{t-i}$ ; $i \in \{1, \ldots, p\}$ is a lagged value of stationary time series (AR-terms) and $\theta_i e_{t-i}$ ; $i \in \{1, \ldots, p\}$ the lagged error (MA-terms). The resulting forecast is yielded by reverse differencing:

$$\hat{Y}_t = \hat{y}_t - Y_{t-1} \qquad (7)$$

The denotation of the yielded model is $ARIMA(p, d, q)$, where $p$ is AR-term, $d$ is number of differencing done and $q$ is the MA-term. There is a systematic procedure for determining the values these parameters should have, which is based on analyzing the autocorrelations (AFCs) and partial autocorrelations (PAFCs) of $y$. The autocorrelation of $y$ at lag $k$ is the correlation between $y_t$ and $y_{t-k}$. The partial autocorrelation of $y$ at lag $i$ is the amount of correlation between $y_t$ and $y_{t-i}$ that is not already explained by the fact that $y_t$ is correlated with $y_{t-1}$, $y_{t-1}$ is correlated with $y_{t-2}$, and the rest of the chained correlation steps until the step $corr(y_{t-i-1}, y_{t-i})$. For the detailed review on the subject, there are several lecture notes and reference books on the subject [23–24].

## 2.3 Bagging method with regression trees

In the field of supervised learning, a regression algorithm is employed to learn the approximation function from the input variable $x$ to the output variable $y$. In the following, for data consisting of $p$ inputs and an output for each of the $N$ observations, the denotation for the input-output pair is $(x_i, y_i)$ for $i = 1, 2, \ldots, N$, with $x_i = (x_{i1}, x_{i2}, \ldots, x_{ip})$.

For the bootstrap aggregation method with regression tree as the base estimator, we consider first the regression tree. A regression tree partitions the feature space into regions, and then fits a simple model in each one. That is,

$$\hat{f}(X) = \sum_{m=1}^{N_m} c_m I\{(X_1, X_2) \in R_m\} \qquad (8)$$

where $Y = \hat{f}(X)$; $x_i \in X$ is the resulted prediction, $N_m$ number of all regions, $c_m$ a constant, $I$ the impunity measure and $R_m$ set of all regions; these terms are defined in more detail in the following. The tree is constructed by applying binary splitting to the data and then evaluating each value of each attribute in the data in order to minimize the cost criterion. That is, in a region $R_m$ the algorithm finds the constant estimators $\hat{c}_1$ and $\hat{c}_2$ greedily in a pair of half planes $R_1$ and $R_2$ by splitting a variable $j$ at split point $s$:

$$R_1(j, s) = \{X | X_j \leq s\}; \; R_2(j, s) = \{X | X_j > s\} \qquad (9)$$

The algorithm then seeks to minimize the mean squared error (MSE) for each of these half-planes:

$$\min_{j,s}[\min_{c_1} \sum_{x_i \in R_1(j,s)}(y_i - \hat{c}_1)^2 + \min_{c_2} \sum_{x_i \in 2(j,s)}(y_i - \hat{c}_2)^2] \qquad (10)$$

where

$$\hat{c}_i = ave(y_i | x_i \in R_i(j, s)); \; i=\{1,2\} \qquad (11)$$

The splitting process is continued until the defined depth, which in our case is until the leaves, i.e. until regions of a size 1. The resulted tree is then pruned using cost-complexity as a criterion. That is, the algorithm prunes the region $R_m$ until that produces the subtree $T$ with minimized cost-complexity $C_\alpha(T)$:

$$C_\alpha(T) = \sum_{m=1}^{|T|} N_m Q_m(T) + \alpha |T|, \quad (12)$$

where $\alpha$ is a tuning parameter (in this case, for full tree, it is 0), $|T|$ is the number of terminal nodes, $N_m$ is number of all regions, $Q_m(T)$ the impurity measure defined by MSE for each region. That is,

$$Q_m(T) = \frac{1}{N_m} \sum_{x_i \in R_m} (y_i - \hat{c}_m)^2. \quad (13)$$

The algorithm continues until it produces the single-node (root) tree, which in the end results in a set of subtrees containing the unique subtree minimizing the cost-complexity for the tree (Eq. 12).

Furthermore, the decision tree is used as the base estimator for bootstrap aggregation (or bagging) that takes multiple samples from our training data with replacement and then trains a model for each random sample. It averages the predictions of all of the sub-models, and then aggregate these averaged predictions to form a final prediction. This is done in order to reduce the variance in our model, as small changes in the data can result in a very different series of splits if only the base estimator were used. [25–27].

### 2.4 Evaluation of methods

We evaluate our methods using two different metrics, which reflect their applicability to quantify the accuracy in the practical application; the first is the root-mean-squared error (RMSE) in the SoH estimation. RMSE is selected as the loss function as it suits to regression model evaluation, and consequently to time series forecast model evaluation [9]. The second is $EVAR$ for the supervised learning methods.

For $RMSE_{SoH}$, all values of the SoH until the last observation and estimator pair and are used to determine its value.

$$RMSE_{SoH}(\hat{y}_i, y_i) = \sqrt{\frac{1}{N} \sum_{i=0}^{N} (\hat{y}_i, y_i)^2} \quad (14)$$

where $\hat{y}$ is the estimator and $y$ the (correct) target output.

Furthermore, we evaluate with the Wilcoxon signed-rank test the normalized ARIMA time series; the Wilcoxon signed-rank null hypothesis (H0) tests if two related paired samples come from the same distribution or not. The test results can be used to assess if a uniform forecast model can be applied to the set of batteries.

For EVAR, used for scoring the supervised learning algorithms, we applied cross-validation by splitting the dataset into 5 parts, so called folds. The algorithm is trained on $k-1$ folds (here: 4-folds) with one held back, and the algorithms is tested on the held back fold. This is repeated so that each fold is held back as test set; the result is yielded as the mean and standard deviation of the combined scores. Rather than using just a single arbitrary value of SoH, this attains more accurate estimate for the new data. The underlying $EVAR$ scoring is yielded by the following equation:

$$exp_{var}(y, \hat{y}) = \frac{Var\{y - \hat{y}\}}{Var\{y\}}, \quad (15)$$

where $\hat{y}$ is the estimator and $y$ the (correct) target output. The best possible score is 1.0; lower values are worse.

Finally, the bagging model was compared to the following suite of three non-linear and ensemble methods: CART is a parametric model and the rest are non-parametric models. Parametric models summarize the data with a set of parameters and requires restricted hypothesis to avoid overfitting. Non-parametric models cannot be characterized with a set of parameters as they derive results from the set of new data with various methods [15]. Finally, it can be noted that the reasoning between connection of the data input and the yielded result (here SoH) is not simple, and often not done for non-parametric models.

## 3. RESULTS AND DISCUSSION

### 3.1 Parameter models – results

For verifying the selected SoH calculation method, we selected five batteries' (A–E) time series randomly from a suite of 31 batteries. Data plot of battery SoH shown as monthly mean values for the batteries over 32 monthly observations (Fig. 2).

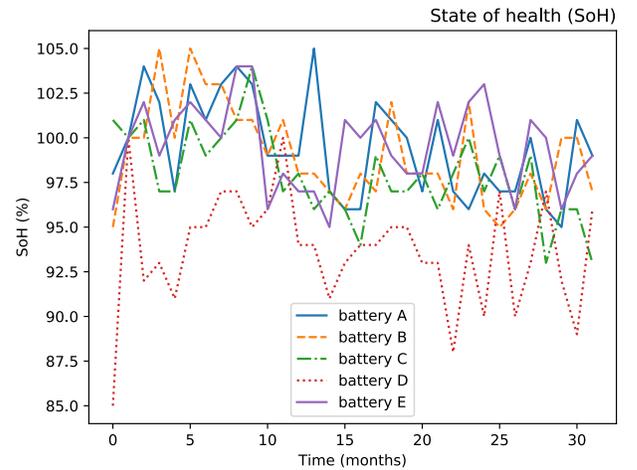

**FIGURE 2:** SoH of five batteries over 32 months.

The data used for forecasting were the monthly SoH average values as this streamlines out some sparse outliers in the data that were still left there due to transient failures. In the resulting monthly averaged data (Fig. 2), there are no obvious outliers; however, there is remarkable fluctuation month-to-month, up and down. The battery A SoH-curve represents a typical behavior with a slight upward trend in the very beginning of the data series and downward trend later on. Small increases in capacity after a slow cycle or rest period may result in SoH exceeding 100% [11]. This was visually confirmed from plots of 31 Li-ion batteries for this work. Therefore, the battery A was selected for developing the forecast model.

The modelled full charging cycles have a linear trend (Fig. 3). Extrapolating from the cycle count graph, if the usage behavior remains unchanged, the designed model estimates the truck's battery pack to complete around 3000 cycles in a ten-year service period. Estimated cycle life corresponds to the published cycle life for commercial Nickel-Cobalt-Manganese (NMC) cells [1].

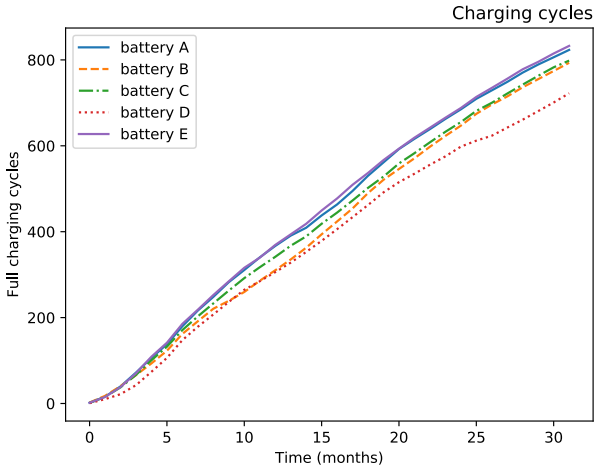

**FIGURE 3:** Equivalent Cycles for batteries A, B, C, D and E.

### 3.2 ARIMA model -results

The first forecast example that we describe is forecasting SoH by ARIMA. We employed the basic non-seasonal ARIMA, although there was an argument for adding the seasonal part, as the ambient temperature chart indicated (Fig. 1); however, for the reasons explained in the following, we selected basic ARIMA for model development. Most importantly, the seasonality is not anymore that clear for the SoH (Fig. 2, 6) itself.

In order to identify a suitable ARIMA-model, we developed a baseline for the forecast modelling by assessing first the naïve forecast (also called persistence) for battery A. That is, the time series was split to two: persistence train and test sets. In the persistence forecasting the observation from the previous step was used as the prediction to the observation on the next step. This naïve forecast's loss function $RMSE_{SoH}$ yielded 3.37 for battery A.

To highlight the importance of the underlying data for developing the ARIMA model, as the next step in the model development, a histogram and density plot of the SoH observations was made without taking to account of any temporal structure in this data set. It shows that the distribution is not Gaussian, and it is right shifted (Fig. 4).

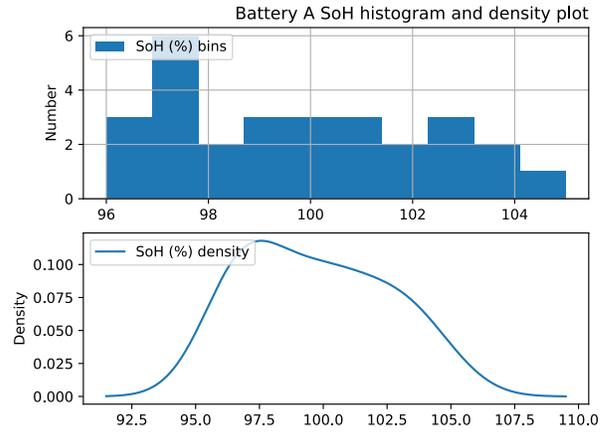

**FIGURE 4:** Plots of SoH of battery A over 32 months.

In the following step, we made a yearly box-and-whisker plot on battery SoH. It indicates a 2.2% yearly depreciation for SoH (Fig. 5). Linear extrapolation from this 2.2% value yields a time span of 9–10 years before the SoH reaches 80% level. The 80% is commonly used SoH reference value for the end of usable lifetime for a truck battery, see for example [1]. This also corresponds with the extrapolated cycle life values (3000 cycles in 9.5 years) calculated above (see Eq. 3 and Fig. 3). Further, the factory has promised 8 years of lifetime for these batteries; therefore, this finding agreed with the promised value taken the safety margin into account.

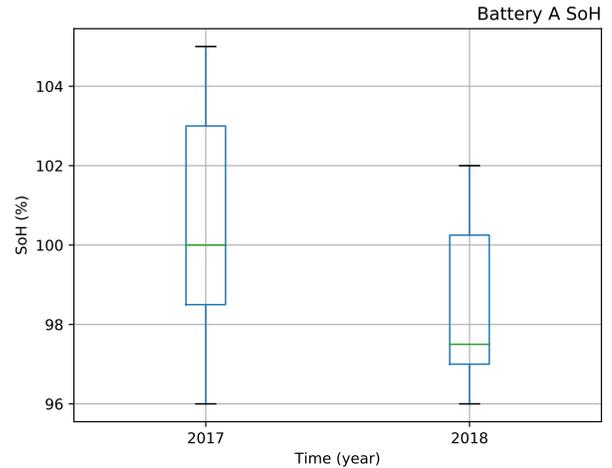

**FIGURE 5:** Battery A SoH degradation over 2 years.

Followingly, year-on-year view was used to verify the hypothesis of seasonality. As illustrated in the charts, some seasonality is seen e.g. around August, i.e. holiday and hot season in the northern hemisphere (Fig. 1). Also, some seasonality seems to be present for SoH around November (Fig. 6). However, the time series is too short to develop seasonal ARIMA model and at the same time hold back a reasonable long verification part of the time series. Furthermore, this phenomenon was not clear comparing the end of years 2017 and 2018 for SoH (Fig. 6).

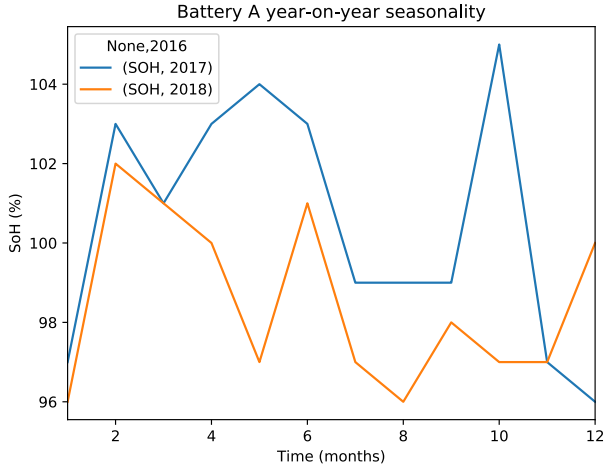

**FIGURE 6:** SoH year-on-year seasonality of battery A.

However, the time series was not stationary; therefore, we made a first order differentiation (Eq. 5) for the time series of battery A to make it stationary. For the resulted time series, $y$, the augmented Dickey-Fuller test (ADF) yielded test statistic value -5.4 that is smaller than the critical value -3.7 at 99% confidence level. i.e. the 1-lag differenced time series did not have time-dependency anymore. Hence the ARIMA model d-value was set to at least 1.

In the next step, the $ARIMA(p,d,q)$ parameter values were selected with the help of autocorrelation function (ACF) and partial autocorrelation function (PACF) analysis with the help of reference [18]. A quick analysis of the ARIMA(0,1,1) yielded the second best $RMSE_{SoH}$ of 2.95. The ARIMA(0,1,1) was an exponential smoothing; an average of the last few observations in order to filter out the noise and more accurately estimate the local mean by exponentially weighted average. [25].

Ideally the model's residual errors are Gaussian, and the mean is zero. The results indicated at least some Gaussian like distribution for ARIMA (0,1,1) (Fig. 7).

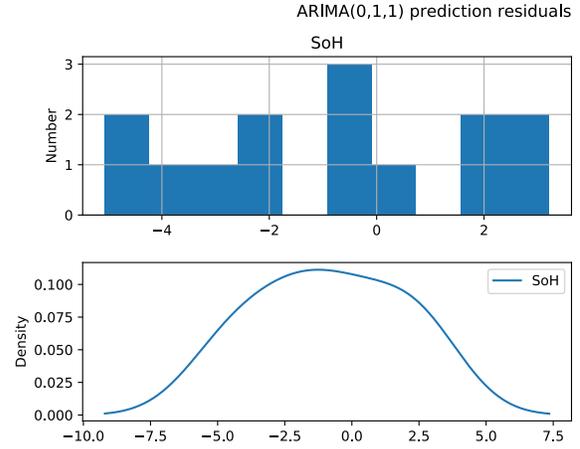

**FIGURE 7:** Plots of SoH of battery A with ARIMA(0,1,1) model.

Having selected the ARIMA model, we investigated its performance for forecasting. Firstly, the selected model makes the assumption that the SoH linearly decreases. With this assumption in mind, we forecasted in a rolling-forecast manner. That is, the prediction algorithm stepped over lead times in the validation dataset and took the observations as an update to the history (i.e. training data).

The first forecast yielded SoH value of 97.5 vs. 96.0 in the validation data set (Fig. 8a–b). We noted that the prediction was around the observed value, and as the next step, the set of predictions was made. In the end, this yielded $RMSE_{SoH}$ of 2.68, which was better than e.g. the baseline naïve forecast result of 3.37. This RMSE result corresponds to the published predictions with similar kind of prediction horizon [7].

The goodness of the fit for ARIMA, $R^2$, is better than for naïve model, albeit the yielded result of -0.26 is not promising. It can be said that the found ARIMA model doesn't fit well; however, this may be due to the relatively short (32 months) time series for developing the ARIMA model. Furthermore, it can be noted that as the assessment of the overall RMSE value reflects comparable results in the field of battery forecasting, hence this realization of $R^2$ may not reflect the intended use of the model and model quantities in relevant manner [28].

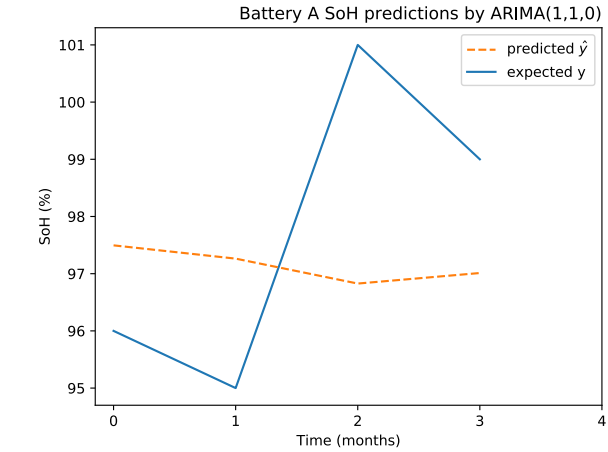

(a)

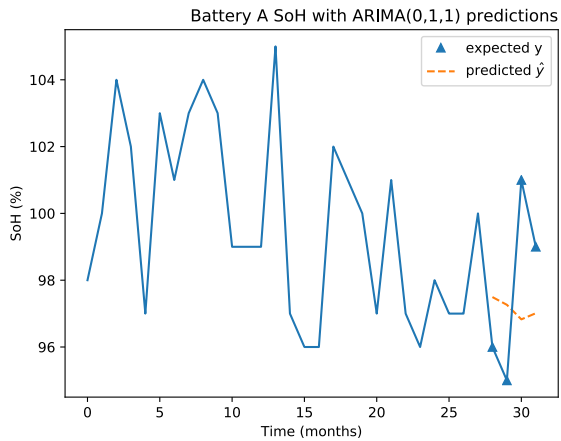

(b)

**FIGURE 8:** Battery A **(a)** ARIMA (0,1,1) predicted SoH values vs. the test set expected values and **(b)** all SoH values together with ARIMA (0,1,1) predicted SoH values.

As both predicted and expected values are around the same, we made the Student t test to see if the means of the two samples (train and test sets) were the same. The Student t -test on paired samples (orange and blue in Fig. 8a–b) yielded that the sample means are equal. This indicates that the model is capable of making prediction of SoH values to some extent.

### 3.3 Comparison of 31 battery time series -results

As we have batteries from the same factory that have the same calendar age, we may expect the SoH trends to be correlated. Therefore, from the population of 31 batteries, we tested the batteries that had similar length of raw observations (32 months) in order to verify if they represented similar kind of sample distributions. Here we made assumptions that the difference in operating temperatures or driving patterns may have significant impact on the individual battery SoH behavior [29–30]. Finally, there were 14 batteries that had 32 months of data including the battery A; the rest of the time series were shorter.

We normalized the data (Eq. 5) in order to make the Wilcoxon significance test. The hypothesis (H0) was set followingly: the sample distributions from different batteries were related to the battery A. Wilcoxon yielded that 100% of the batteries had same distribution than battery A (failed to reject H0), and consequently none had different distribution (rejected H0). Hence, we have clear indication that this ARIMA(0,1,1) is among the best algorithm for the set of batteries.

It is promising that some of the batteries are within the same distribution; however, there is a need for further analysis on e.g. the environmental factors on the battery SoH forecast and the amount of data for relevant time series forecasting when using this method.

### 3.4 Bagging with decision trees -results

In the following example that we describe forecasting SoH by bootstrap aggregation (bagging) using decision trees as the base estimator. In order to improve the forecast results, we calculated the correlations of all recorded time series features (Fig. 9). Strongly correlated data tends to bias the resulted model, and the model may benefit from obviating some excess features. However, some machine learning algorithms, can mitigate this problem themselves to some extent. In practice, they may benefit from feature reduction.

Because of this, we made principal component analysis and feature importance analysis by extra trees classifier. These suggested that at least energy, charging pulse frequency, and ambient temperature were significant features for predicting the SoH.

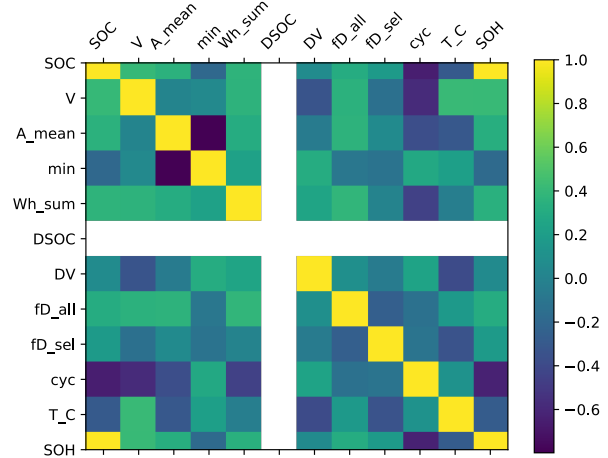

**FIGURE 9:** Battery A feature correlation matrix.

For verifying the set of used features, all selected supervised learning methods were scrutinized with different set of features to find the best feature set for the train set. Using all 11+1 features (here +1 is the target SoH) yielded *EVAR* scores of 0.85–0.93, (Eq. 15); due to the statistical nature of the algorithms the exact results vary each execution time a bit. Selecting the following features: SOC, charging pulse mean current, charging pulse minutes and charging pulse frequency yielded *EVAR* scoring 0.93-0.95, which is better than using all features.

Finally using only two features, SOC and charging pulse frequency, the *EVAR* yielded 0.94–0.97; BAG, RF and GBM were on par (Fig. 10). However, bagging was chosen as it has tolerance to variance in the data [26].

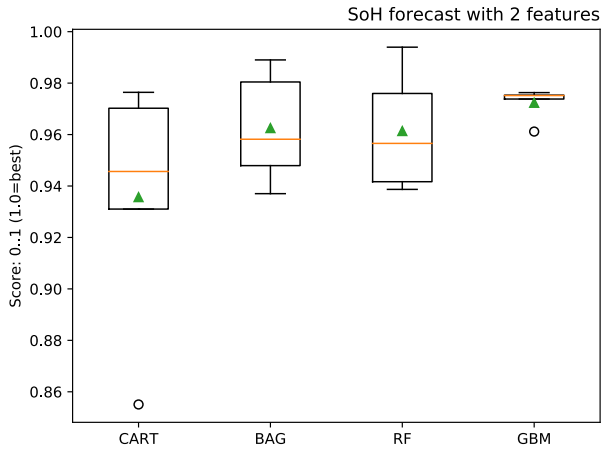

**FIGURE 10:** Comparing algorithms using 2 features.

For making a final comparison with the ARIMA model, the feature distributions were analyzed. There was no Gaussian distribution, and the feature scales are different; hence, the battery data was standardized for the best-found regression algorithm, i.e. bagging. Albeit a difference to ARIMA model is that the bagging method does not necessarily require feature scaling to yield reasonable results, the feature scaling seemed to help its performance in this case. After standardizing the raw data, grid search found the best hyperparameters for the finalized BAG model. This finalized model yielded *EVAR* score over 0.95, which was a satisfactorily result.

Having selected the bagging model, we investigated its performance for forecasting in walk-forward manner, adding a forecast at one step and evaluating the results to the test set of SoH values that were hold back; the first predicted SoH value was 96.08 vs. expected 96.00 in the validation data set (Fig. 12a-b).

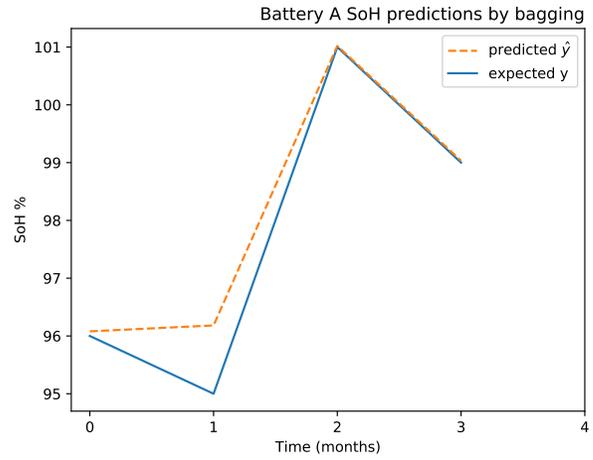

(a)

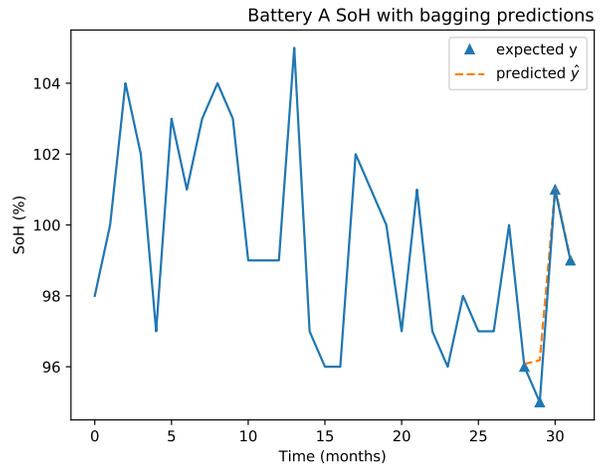

(b)

**FIGURE 11:** Battery A **(a)** BAG predicted SoH values vs. the test set expected values, and **(b)** all observed values and the BAG predicted SoH values.

For BAG, the loss function $RMSE_{SoH}$ yielded 0.59 (%), which was much better than the naïve model's $RMSE_{SoH}$ of 3.12 (%). The model was skillful in comparison to other models.

The goodness of the fit, $R^2$ score, is fairly good (0.94) for the final BAG model (Table 2). , This corresponds to the previously yielded fairly good *EVAR* score (0.95).

Lastly, it should be noted that the BAG model can be overconfident in its predictions indicated by this $R^2$ value that is close to one.

| Method/scoring | $RMSE_{SoH}$ | $R^2$ |
|---|---|---|
| Persistence | 3.37 | -1.72 |
| ARIMA(0,1,1) | 2.68 | -0.26 |
| BAG-normalised | 0.59 | 0.94 |

**TABLE 2:** Summary of model evaluations.

## 3.5 Discussion

In the first part, we demonstrated that the ARIMA (0,1,1) model is applicable to the battery SoH forecasting; however, the resulted loss function $RMSE_{SoH}$ value indicated that the model has room for enhancing. Furthermore, we have shown that for a set of batteries Wilcoxon test found that some of the batteries come from the same distribution; this suggests that a common forecast model can be developed for the set of batteries. However, there are indicators that the environmental conditions need to be taken into account better when developing the model further. Lastly, we showed that the developed bagging model achieves superior forecasting accuracy.

As the time series are relatively short (32 months) and as they may show seasonality in the long run, this in turn indicates a drawback in our models. The availability of more data sets with relatively long time series will be necessary to resolve issues on the forecasting accuracy, which is a known issue in the field of batteries [31]. In the case of ARIMA(0,1,1), it assumes linear trend when in reality it may be non-linear.

The drawback for non-linear bagging (BAG) model is a tendency to overfit; therefore, this forecast model's loss function may start to grow from the original one if the data input changes to e.g. non-linear one later on. Overall, the set of 32-month battery time series provided us a foundation for the model development work; however, in the context of 10+ years of expected life-time for lithium-ion batteries, the results indicate that more observations would benefit to enhance forecast accuracy of both of these models for forecasting SoH.

## 4 CONCLUSIONS

This paper has demonstrated the applicability of ARIMA and supervised learning methods for battery state of health (SoH) forecasting under circumstances when there is little prior information available about the batteries. A supervised learning method, bootstrap aggregation (bagging) with decision trees is proposed as the base regression estimator, for making forecasts and the results are validated by comparison with other regression models. It demonstrates how supervised-learning-enabled SoH prognosis can effectively exploit data from multiple cells in lithium-ion batteries from 31 EV forklifts to significantly improve the forecasting performance. The main bottleneck in the approach is relatively short (32 months) time series data which may show seasonality, or other changes due to the battery chemistry in the long run.

The future work could be the consideration of driver behaviors, temperature and SOC dependence to further examine the multivariate SoH predictor.

## ACKNOWLEDGEMENTS

This work has been supported by the European Commission through the H2020 project FINEST TWINS (grant No. 856602) and by the Academy of Finland through the project PUBLIC (grant No. 322742). The authors appreciate the invaluable support of Professor Tanja Kallio and Mr. Panu Sainio from Aalto University in helping to evaluate the data and develop the research direction.

This paper has first been published at the Proceedings of the ASME 2020 International Mechanical Engineering Congress and Exposition IMECE2020 held at November 16-19, 2020, Portland, OR, USA (paper ID 23949).